\documentclass[letterpaper, 10 pt, conference]{ieeeconf} 

\IEEEoverridecommandlockouts                             

\usepackage{comment}
\usepackage{graphicx} 
\usepackage{amsmath} 
\usepackage{amssymb}  
\usepackage{cite}
\usepackage{xcolor}
\usepackage[hidelinks]{hyperref}
\usepackage[caption=false, font=scriptsize, labelfont=sf, textfont=sf]{subfig}
\usepackage{bm}
\usepackage{adjustbox}
\usepackage{placeins}
\usepackage{soul}
\usepackage[font={small}]{caption}

\title{\LARGE \bf

Impact-Friendly Object Catching at Non-Zero Velocity Based on Combined Optimization and Learning
}

\author{Jianzhuang Zhao$^{1,2}$, Gustavo J. G. Lahr$^{1}$, Francesco Tassi$^{1,2}$, 
Alessandro Santopaolo$^{1}$, \\
Elena De Momi$^{2}$, and Arash Ajoudani$^{1}$
\thanks{$^1$ Human-Robot Interfaces and Interaction Lab, Istituto Italiano di Tecnologia, Genoa, Italy. \tt\small jianzhuang.zhao@iit.it}
\thanks{$^2$ Dept. of Electronics, Information, and Bioengineering, Politecnico di Milano, Italy.}
\thanks{This work was supported by the European Research Council's (ERC) starting grant Ergo-Lean (GA 850932).}
}

\begin{document}

\maketitle
\thispagestyle{empty}
\pagestyle{empty}

\begin{abstract}
This paper proposes a combined optimization and learning method for impact-friendly, non-prehensile catching of objects at non-zero velocity. Through a constrained Quadratic Programming problem, the method generates optimal trajectories up to the contact point between the robot and the object to minimize their relative velocity and reduce the impact forces. Next, the generated trajectories are updated by Kernelized Movement Primitives, which are based on human catching demonstrations to ensure a smooth transition around the catching point. In addition, the learned human variable stiffness (HVS) is sent to the robot's Cartesian impedance controller to absorb the post-impact forces and stabilize the catching position. Three experiments are conducted to compare our method with and without HVS against a fixed-position impedance controller (FP-IC). The results showed that the proposed methods outperform the FP-IC while adding HVS yields better results for absorbing the post-impact forces. 
\end{abstract}

\section{Introduction}

Dynamic robotic manipulation tasks are challenging since they require tight coordination between object detection, motion planning, and control. Activities such as throwing~\cite{tossingbot2020tro}, hitting~\cite{tennis2020iros}, juggling~\cite{juggling2021iros}, and catching~\cite{hardcat2014tro,softcat2016tro} play an important role in various applications, including logistics and aerospace. 
Catching objects is particularly challenging as the overall catching motion demands fast and precise planning with quick execution. 
Moreover, the impact force between a robot and a flying or falling object during the contact phase should be reduced to avoid causing damage to both sides.

\begin{figure}[t]
    \centering
    \includegraphics[trim=0.0cm 0.2cm 0.0cm 0.15cm,clip,width=0.9\columnwidth]{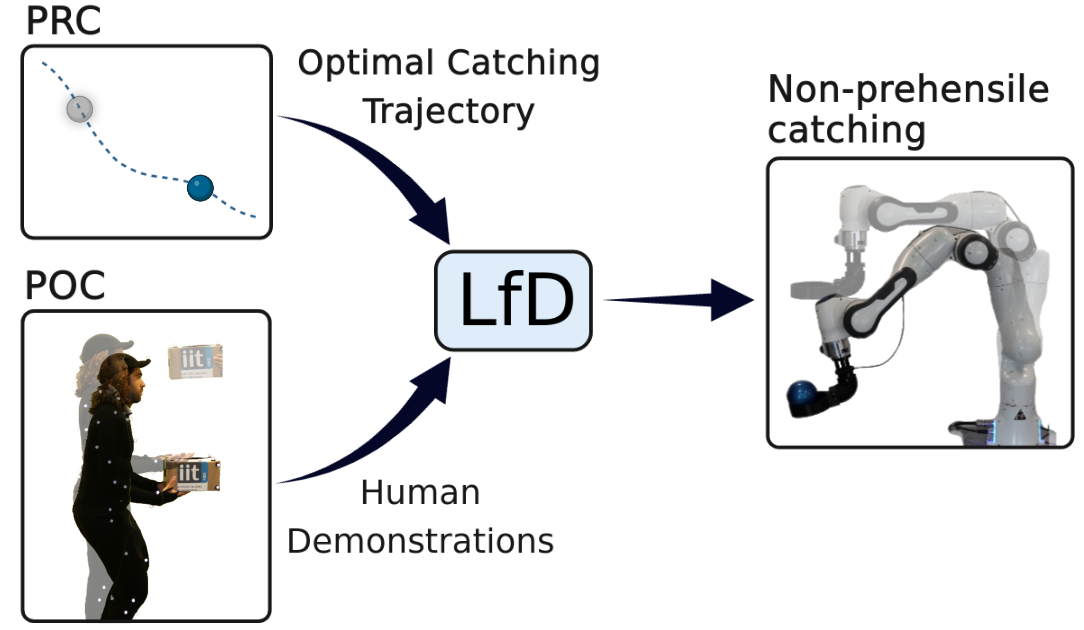} 
    \caption{The robot achieves impact-friendly and safe non-prehensile catching behavior through optimization in PRC, to reduce the impact force, and learn from human demonstrations in POC, to absorb the post-impact force and avoid bouncing and second impact.}
    \label{fig:con_fig}
    \vspace{-5mm}
\end{figure}

 Based on the end-effector type, catching can be divided into closed-form (i.e., multi-fingered hand) and non-prehensile catching (Fig. \ref{fig:con_fig}). In the former case, the object cannot be dropped once it is caught and the fingers are tightly closed. For the latter, with a high-stiffness controlled robot, the object may bounce away and lose contact, resulting in a second impact~\cite{ajoudani2012tele}, because of the post-impact force. Therefore, for non-prehensile catching tasks, the impact forces should be reduced, and the post-impact forces need to be absorbed.

\begin{figure*}[t]
	\centering
	\includegraphics[trim=0cm 0cm 0cm 0.85cm,clip,width=.95\linewidth]{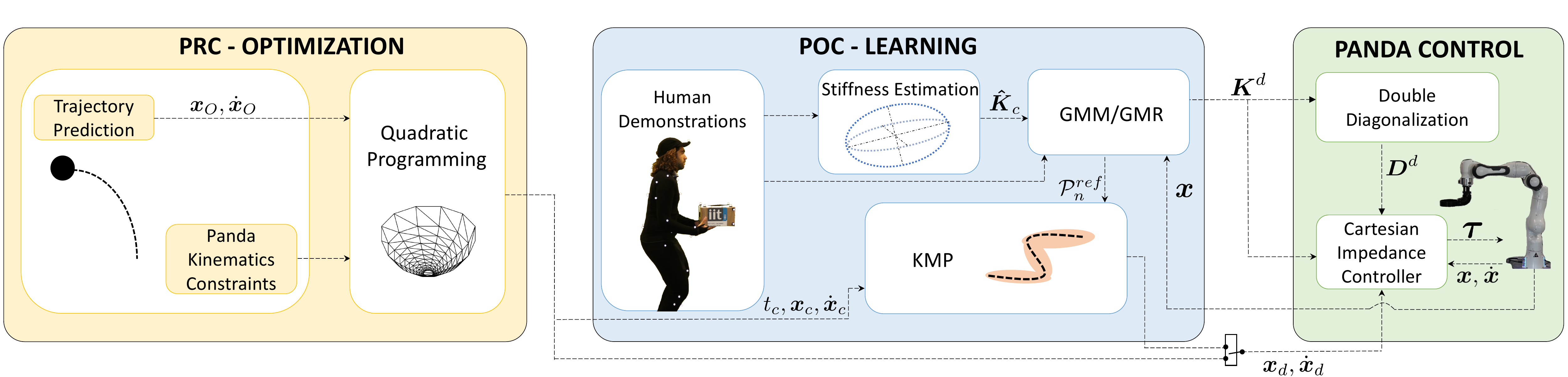}
	\caption{Overall process of the proposed framework. From left to right, the PRC optimization receives the initial object's pose as input and computes the PRC optimal catching trajectories via constrained QP. For the POC phase, human demonstrations are used to obtain the post-impact trajectories via KMP and to learn the variable stiffness profile used in the VIC. Lastly, the lower-level Cartesian impedance controller generates the actuation torques provided to the robot. After the PRC phase, the POC starts publishing the learned desired trajectory and variable impedance profiles.}
	\label{fig:framework}
	\vspace{-5mm}
\end{figure*}

The catching motion can be divided into two phases: Pre-Catching (PRC) and Post-Catching (POC). PRC represents the period before the first contact between an object and a robot. After a successful PRC, the POC stage starts and is defined by the period following the impact until the complete movement stoppage. A successful POC phase should dissipate the energy the impact injects without overloading the robot's internal torques and avoiding bouncing as much as possible. As impedance controllers are regularly used for interaction in POC, it is worth pointing out that their stiffness does not affect the impact force due to the extremely short impact period (a few milliseconds ~\cite{wangmodel22}). Still, they are fundamental to help to dissipate the post-impact forces~\cite{haddadin2009requirements,ajoudani2012tele}, prevent damaging the object \cite{Uchiyama2012egg} and the robot, especially in tasks with high speed \cite{9636775} and/or high object mass \cite{stouraitis2020multiHybrid}.

Even though PRC and POC are isolated from a modeling perspective, the success of POC strongly relies on the performance of PRC since the impact force mainly depends on the relative robot-object velocity and their reflected inertia \cite{stouraitis2020multiHybrid, Uchiyama2012egg, tassi2022,wang2019impactrss}.
Velocity matching (VM) is often employed to minimize the relative velocity by planning the relative motion between an object and the robot. A good VM performance in PRC depends on perception and planning strategies combined with fast online trajectory planning/re-planning and control for manipulators. Several studies proposed solutions for VM with online re-planning for catching, using model-based polynomial \cite{ poly2015brun} or optimization \cite{opball2011jan} strategies. Although these methods generate smooth trajectories, they do not consider VM's effect on the efforts caused by the impact during POC. 

Other solutions for VM take inspiration from humans. Learning from (human) demonstrations (LfD) approaches are powerful techniques to teach a robot to produce fast and reactive motion. Humans can catch moving objects with non-zero velocity, generating fast yet smooth trajectories, even for unforeseen objects~\cite{kajikawa1999analysis}. In \cite{hardcat2014tro}, the robot motion is learned from human throwing demonstrations, but the robot stops directly when the objects are contacted with the EE (multi-fingered hand), which may result in the bouncing of the object. To address this problem, in \cite{softcat2016tro} the robot continues to track the predicted objects' path after contact, generating a soft-catching motion. 
LfD studies have generated promising results, yet they did not consider the post-impact dynamics during POC, limited to catching lightweight objects. 

Therefore, designing robot controllers with compliant behavior is paramount to dealing with uncertainties, either in tracking or the object's properties, and maintaining stability after contact. It is possible to obtain highly damped behaviors using impedance \cite{Senoo2016plasticImpedance} or admittance \cite{Fu2021admittance} controllers, but the controller's parameters are often constant, and their choice is still mainly empirical. In \cite{stouraitis2020multiHybrid}, a hybrid controller was implemented with multi-mode trajectory optimization to halt a heavy object traveling on a table at a constant speed of $0.88m/s$. However, due to the transition from free to coupled motion, the switching between different modes results in high nonlinearities, and initializing the whole system is nontrivial. Furthermore, this method is unsuitable for catching flying objects, given their high velocity and accelerating nature.

Similarly to VM in PRC, the control principles during the POC phase can also benefit from inspiration from humans. Tele-impedance was proposed in our previous work~\cite{ajoudani2012tele}, where EMGs estimate the human upper arm stiffness in real-time by remotely controlling a slave robotic arm in a falling ball-catching experiment. 
The results show that the variable impedance control (VIC) outperforms the constant one under several metrics. LfD setups have also been used to teach impedance profiles in quasi-static tasks \cite{abu2018forceras, Zhao2022}. Specifically for catching,  in \cite{phung2014learning} the authors encoded the kinematic human motion with a Gaussian mixture model (GMM) and Gaussian mixture regression (GMR) to catch a flying ball. However, no impact forces are reported to evaluate the method, and no information about human dynamics is encoded to improve the method. Transferring humans' compliant behavior to autonomous robots for catching flying or falling objects is still an open issue. 

In conclusion, in most catching applications, it is necessary to deal with high impact forces, which can affect the task's success and potentially damage the robot. In this work, we employ a robot with a small payload compared to the forces generated during such a dynamic task, intending to overcome these limitations through the proposed strategy.

To address these issues, we propose a method to generate an impact-friendly and safe, dynamic falling non-prehensile object-catching behavior for autonomous robots. Fig. \ref{fig:con_fig} shows the overall approach: a model-based quadratic programming (QP) during PRC to leverage VM, combined with LfD, that generates kinematic and dynamic trajectories during POC. Thus, the contributions are threefold: a novel framework capable of dealing with PRC and POC; a new constrained QP problem is designed to find the optimal catching point, reducing the impact force; a learned VIC is presented based on human kinematic and impedance profiles, which is triggered during the impact to absorb the post-impact forces and ensure stable contact. \textcolor{black}{The proposed method is evaluated by catching a free-falling object, and the results show that this possible to generate safer and more stable catching behavior compared to the other baselines.}

\section{Methodology}\label{sec:methodology}

Our proposed method deals with the planning and control challenges in both PRC and POC, where the processing units and interconnections are illustrated in Fig. \ref{fig:framework}.
First, we develop a rigid body impact formulation to investigate the interaction force details during the collision (Section \ref{subsec:impact_model}). To yield the desired performance for PRC (Fig. \ref{fig:framework}-left), we implement a VM strategy to reduce the difference between the two velocity vectors and minimize the impact force between the falling object and the robot's EE. A novel model-based QP formulation is introduced (Section \ref{subsec:QP_formulation}) to optimize the catching trajectory in PRC for maximum VM to the extent possible (e.g., depending on the robot's limits).

Next, the POC phase starts. A new planner is triggered based on human demonstrations (center block in Fig. \ref{fig:framework}) to obtain a compliant movement to minimize the post-impact forces and guarantee a damped catch effectively. We use GMM/GMR to get a human-like trajectory (i.e., the reference trajectory $\mathcal{P}_n^{ref}$) for the robot's position and velocity, with time as the input variable and robot trajectories as the output. Since the PRC only generates the trajectories up to the catching point, a new planner is needed in POC to connect the PRC and the learned human's motion. Considering the reasonable generalization and scalability, the Kernelized Movement Primitives (KMP) (Section \ref{subsec:kmp}) is chosen in this paper. Specifically, the KMP updates the POC trajectory based on $\mathcal{P}_n^{ref}$ and on the trajectory planned in PRC, connecting both through the catching point ($t_c,\bm{x}_c,\dot{\bm{x}}_c$) to ensure a smooth transition between the two phases. 
The overall time-dependent desired catching trajectory is generated online, which includes the VM in PRC and the KMP updates in POC.

Finally, the human demonstrations are used to learn the stiffness profiles encoded by the GMM/GMR (Section \ref{subsec:human-arm}), leveraging on human-compliant behavior suitable for energy dissipation. It is worth mentioning that the stiffness profile for POC is decoupled from the trajectory, depending on the real-time EE pose, and is triggered by the contact force since the actual catching point might differ due to tracking inaccuracies. We use a Cartesian impedance control for compliance during the task (Section \ref{subsec:impedance_controller} and right block in Fig. \ref{fig:framework}). Details about human demonstrations can be found in Section \ref{subsec:human-demo}. 

\subsection{Impact Model}
\label{subsec:impact_model}

The robotic manipulator rigid body model in joint space may be written as
\begin{equation} \label{eq:rigid_body_full_eq}
    \boldsymbol{M}(\boldsymbol{q}) \ddot{\boldsymbol{q}} + \boldsymbol{C}(\boldsymbol{q},\dot{\boldsymbol{q}})\dot{\boldsymbol{q}} + \boldsymbol{g}(\boldsymbol{q}) = \boldsymbol{\tau} + \boldsymbol{\tau}_{ext},
\end{equation}
where $\boldsymbol{q}, \dot{\boldsymbol{q}}, \ddot{\boldsymbol{q}} \in \mathbb{R}^n $ are the position, velocity, and acceleration vectors in joint space, respectively, and $n$ is the number of degrees of freedom (DoF). $\boldsymbol{M}(\boldsymbol{q}) \in \mathbb{R}^{n \times n}$ is the joint space inertia matrix, $\boldsymbol{C}(\boldsymbol{q}, \dot{\boldsymbol{q}}) \in \mathbb{R}^{n}$ the Coriolis term, $\boldsymbol{g}(\boldsymbol{q})\in \mathbb{R}^{n}$ the gravity term, and $\boldsymbol{\tau} \in \mathbb{R}^{n}$ is the actuation torques. The external torques generated by the interaction forces are given by $\boldsymbol{\tau}_{ext} = \boldsymbol{J}^T(\boldsymbol{q}) \boldsymbol{F}_{ext}$, being $\boldsymbol{J}^T(\boldsymbol{q}) \in \mathbb{R}^{n \times 6}$ the Jacobian transpose matrix.
The model (\ref{eq:rigid_body_full_eq}) may be rewritten in Cartesian space to facilitate the controller design since the task is defined in the same space \cite{ott2008cartesian}: 
\begin{equation}\label{eq:rigid_body_cartesian}
    \boldsymbol{\Lambda}(\boldsymbol{x}) \ddot{\boldsymbol{x}} + \boldsymbol{\mu}(\boldsymbol{x}, \dot{\boldsymbol{x}}) \dot{\boldsymbol{x}} + \boldsymbol{F}_g(\boldsymbol{x}) = \boldsymbol{F}_{\tau} + \boldsymbol{F}_{ext}.
\end{equation}

All the following definitions are in Cartesian space with respect to the EE: $\boldsymbol{x}, \dot{\boldsymbol{x}},\ddot{\boldsymbol{x}} \in \mathbb{R}^{6}$ are the robot's pose, velocity and acceleration, respectively, $\boldsymbol{\Lambda} = \boldsymbol{J}^{-T}\boldsymbol{M}\boldsymbol{J}^{-1}$ the inertia matrix, $\boldsymbol{\mu}(\boldsymbol{x}, \dot{\boldsymbol{x}}) = \boldsymbol{J}^{-T}(\boldsymbol{C}-\boldsymbol{M}\boldsymbol{J}^{-1}\dot{\boldsymbol{J}})\boldsymbol{J}^{-1}$ the Coriolis, $\boldsymbol{F}_g(\boldsymbol{x})=\boldsymbol{J}^{-T}(\boldsymbol{q})\boldsymbol{g}(\boldsymbol{q})$ the gravity, and $\boldsymbol{F}_{\tau} = \boldsymbol{J}^{-T}(\boldsymbol{q}) \boldsymbol{\tau}$ the actuation forces.

During the PRC and right before the impact, at the moment $t_c$, the robot's velocity is $\dot{\boldsymbol{x}}$ and the object's velocity is $\dot{\boldsymbol{x}}_{o}\in \mathbb{R}^{6}$. Right after the impact, velocities of the robot and the object update to $\dot{\boldsymbol{x}} + \delta\dot{\boldsymbol{x}}$ and $\dot{\boldsymbol{x}}_{o} + \delta\dot{\boldsymbol{x}}_{o}$, respectively. 
The impact phase is defined by the time frame where the control actions are not taking effect. The robot's velocity and configuration will mostly generate reactive behavior.
For an instantaneous collision, given that the direction of contact is defined by the normal vector $\boldsymbol{\eta}$, the coupled dynamics of the robot and the object as rigid bodies are described by:
\begin{equation}
    [(\dot{\boldsymbol{x}} + \delta\dot{\boldsymbol{x}})-(\dot{\boldsymbol{x}}_{o} + \delta\dot{\boldsymbol{x}}_{o})]^T \boldsymbol{\eta} = -e(\dot{\boldsymbol{x}}-\dot{\boldsymbol{x}}_{o})^T \boldsymbol{\eta},
    \label{eq:restitution_coefficient}
\end{equation}
with $0<e<1$ being the coefficient of restitution. When $e=1$ is said to be an elastic collision, the bodies have maximum velocity after the impact; while $e=0$ is a plastic collision, i.e., the relative velocity of the two bodies is zero. 

The interaction forces generate finite impulsive forces $\hat{\boldsymbol{F}}$ at the contact point: $\hat{\boldsymbol{F}}=\lim_{\delta t \rightarrow 0} \int^{t+\delta t}_{t} \boldsymbol{F}_{ext}(s) ds$, where $\delta t$ is the duration of the impact.
From the integration of (\ref{eq:rigid_body_full_eq}), the analysis of impact shows that the variation of velocity in the robotic manipulator due to the impact is given by $\delta \dot{\boldsymbol{x}} = \boldsymbol{\Lambda}^{-1} \hat{\boldsymbol{F}}$ \cite{impactAnalysis2000kim}.
Assuming the object as a point mass (${m}_{o}$) w.r.t. to the manipulator, its velocity variation due to impact is $\delta \dot{\boldsymbol{x}}_{o}= 1/m_{o} (-\hat{\boldsymbol{F}})$ \cite{advancedDynamics2003Greenwood}.

\begin{equation}
    \hat{\boldsymbol{F}} = \left( \boldsymbol{\Lambda}^{-1} + \frac{1}{m_{o}} \boldsymbol{I} \right)^{-1} (\dot{\boldsymbol{x}}_{o}-\dot{\boldsymbol{x}}).
    \label{eq:impact_relationship}
\end{equation}

Equation (\ref{eq:impact_relationship}) shows the dependencies that can minimize the impact forces during catching. The first term depends on the object's mass, which we do not control, and on the robot's inertia, which could be modulated \cite{Haddadin2017inertiaShaping}. Alternatively, the dependency of the inertia from the robot's configuration, via $\bm{J}$, can be exploited \cite{tassi2022}, which will be treated in future works.
The second term shows that VM is essential to minimize the forces, especially in cases where the objects are accelerating at each time step and often will achieve velocities higher than the maximum speed of the robot's EE in short times.
Finally, (\ref{eq:impact_relationship}) works as an upper boundary for deformable objects since a compliant body would passively dissipate energy due to deformation \cite{Siciliano2020DeformableManipulation}.

\subsection{Pre-catching: Online Velocity Matching} \label{sec:prc}
\label{subsec:QP_formulation}

This section evaluates the robot's optimal catching point and catching velocity necessary to achieve online VM, minimizing the impact force, as discussed in \ref{subsec:impact_model}. 
For this purpose, the following QP problem is formulated:
\begin{align}
    \min_{\dot{\bm{x}}_R} \: \frac{1}{2} & \left( \alpha \| 
    \dot{\bm{x}}_R-\dot{\bm{x}}_O \|^2 + \beta \| 
    \bm{x}_R-\bm{x}_O \|^2 - \gamma \| \bm{G} \bm{x}_R \|^2
    \right) \nonumber \\ 
    s.t. \hspace{0.4cm} & \bm{x}_R^{min} \le \bm{x}_{R_{0}}+\Delta t \ \dot{\bm{x}}_R \le \bm{x}_R^{max} \label{eq:general_QP_formulation} \\
    -&\dot{\bm{x}}_R^{max} \le \dot{\bm{x}}_R \le \dot{\bm{x}}_R^{max} \nonumber 
    \\
    -&\Delta t \ \ddot{\bm{x}}_R^{max} \le \dot{\bm{x}}_R-\dot{\bm{x}}_{R_{0}} \le \Delta t \ \ddot{\bm{x}}_R^{max}
    \label{eq:general_QP_formulation} \nonumber 
\end{align}
where $\bm{x}_R,\dot{\bm{x}}_R \in\mathbb{R}^{6 T}$ and $\bm{x}_{R_0}, \dot{\bm{x}}_{R_{0}} \in\mathbb{R}^{6 T}$ are respectively the actual and previous position and velocity trajectories of the robot written in vector form as: $
    {\bm{x}}_R = \begin{bmatrix}
        {\bm{x}}_{{i+1}}^T,
        {\bm{x}}_{{i+2}}^T
        ,..,
        {\bm{x}}_{{i+T}}^T
    \end{bmatrix}^T,
    \dot{\bm{x}}_R = \begin{bmatrix}
        \dot{\bm{x}}_{i}^T,
        \dot{\bm{x}}_{i+1}^T
        ,..,
        \dot{\bm{x}}_{i+T-1}^T
    \end{bmatrix}^T,
    {\bm{x}}_{R_{0}} = \begin{bmatrix}
        {\bm{x}}_{{i}}^T,
        {\bm{x}}_{{i+1}}^T
        ,..,
        {\bm{x}}_{{i+T-1}}^T
    \end{bmatrix}^T,
$
with $\bm{x}_{i}, \dot{\bm{x}}_{{i}} \in\mathbb{R}^{6}$ being the robot's Cartesian poses and velocities respectively at the $i$-th time instant and $T$ is the length of the prediction horizon, chosen based on the time necessary for the object to reach the ground.
$\bm{x}_O,\dot{\bm{x}}_O, \ddot{\bm{x}}_O\in\mathbb{R}^{6 T}$ are the overall object's trajectories as 
$
    {\bm{x}}_O = \begin{bmatrix}
        {\bm{x}}_{o_{i+1}}^T,
        {\bm{x}}_{o_{i+2}}^T
        ,..,
        {\bm{x}}_{o_{i+T}}^T
    \end{bmatrix}^T,
    \dot{\bm{x}}_O = \begin{bmatrix}
        \dot{\bm{x}}_{o_{i}}^T,
        \dot{\bm{x}}_{o_{i+1}}^T
        ,..,
        \dot{\bm{x}}_{o_{i+T-1}}^T
    \end{bmatrix}^T.
$
They are calculated based on the initial height of the object without the need for a separate trajectory estimator, thanks to its simple free-falling kinematics, the short distance, and the ideal conditions.
$\bm{G} = diag \{\left[0,0,1,0,0,0\right],\hdots, \left[0,0,1,0,0,0\right] \} \in\mathbb{R}^{6T \times 6T}$ is a selection matrix responsible for the isolation of the $z-$components of $\bm{x}_R$ (see Fig. \ref{fig:exp_setup} for the reference frame), which is used in the last objective function to maximize the catching height of the object, to counteract the object's constant gravitational acceleration, since the velocity of the object soon exceeds the maximum velocity limits of the robot's EE, thus affecting VM.
$\alpha, \beta, \gamma \in\mathbb{R}$ are the weights associated with each objective function, $\Delta t$ is the planner's sampling period, and $\bm{x}_R^{min},\bm{x}_R^{max}, \dot{\bm{x}}_R^{max}, \ddot{\bm{x}}_R^{max} \in\mathbb{R}^{6 T}$ are the boundaries for minimum and maximum position, velocity, and acceleration, respectively.

The cost function of \eqref{eq:general_QP_formulation} establishes a soft hierarchical order. In particular, the primary objective is to catch the object successfully (thus $\beta > \alpha$), while the secondary objective is to achieve VM. The last term maximizes the catching height to minimize the falling distance traveled from the object, improving VM.
These gains are fine-tuned based on the trade-off between position accuracy and VM required (see section \ref{subsec:exp-set}).
By identifying online the initial position of the object and generating its free-falling trajectory, the QP problem \eqref{eq:general_QP_formulation} finds the optimal position and velocity trajectories $\bm{x}_R,\dot{\bm{x}}_R$ that the robot should follow as references (see Fig. \ref{fig:framework}) until the catching point at time $t_c$, in which the desired pose and velocity of the robot are $\bm{x}_c,\dot{\bm{x}}_c$ and the POC phase is triggered.

\subsection{Post-catching}\label{sec:poc}

\subsubsection{Kernelized Movement Primitives}
\label{subsec:kmp}

KMP~\cite{huang2019kernelized} is a newly proposed imitation learning method from the information theory perspective. A probabilistic reference trajectory $\mathcal{P}_n^{ref} = \mathcal{N}(\boldsymbol{\hat{\mu}}_n,\boldsymbol{\hat{\Sigma}}_n)$ is extracted by GMM/GMR from human catching demonstrations (middle block in Fig. \ref{fig:framework}), where $\boldsymbol{\hat{\mu}}_n$ and $\boldsymbol{\hat{\Sigma}}_n$ represent the mean and covariance, respectively. Then, the derivation of KMP starts from a parametric trajectory
\begin{equation}
	\boldsymbol{\xi}(\boldsymbol{s}) =\boldsymbol{\Phi}^{T}(\boldsymbol{s})\boldsymbol{w},
	\label{eq:kmp}
\end{equation}
where $\boldsymbol{\Phi}(\boldsymbol{s})=\boldsymbol{I}_\mathcal{O}\otimes\boldsymbol{\phi}(\boldsymbol{s})\in\mathbb{R}^{B\mathcal{O}\times\mathcal{O}}$, $\boldsymbol{\phi}(\boldsymbol{s})$ denotes $B$-dimensional basis functions and $\boldsymbol{I}_\mathcal{O}$ is the $\mathcal{O}$ dimensional identity matrix. The weight vector is $\boldsymbol{w}\sim\mathcal{N}(\boldsymbol{{\mu}}_w,\boldsymbol{{\Sigma}}_w)$, where $\boldsymbol{{\mu}}_w$ and $\boldsymbol{{\Sigma}}_w$ are unknown. To obtain these variables, KMP uses \eqref{eq:kl} to minimize the KL-divergence between the probabilistic trajectory generated by \eqref{eq:kmp} and the reference trajectory $\mathcal{P}_n^{ref}$

\begin{equation}\label{eq:kl}
	\sum_{n=1}^{N}{KL}(\mathcal{P}_n^{para}||\mathcal{P}_n^{ref}),
\end{equation}
where $\mathcal{P}_n^{para}=\mathcal{N}(\boldsymbol{\Phi}^{T}(\boldsymbol{s}_n)\boldsymbol{{\mu}}_w,\boldsymbol{\Phi}^{T}(\boldsymbol{s}_n)\boldsymbol{\Sigma}_w\boldsymbol{\Phi}(\boldsymbol{s}_n))$. By decomposing the above objective function, for any input $s^*$, the corresponding output mean and covariance are computed as:

\begin{equation}\label{eq:klmean}
\mathbb{E}(\boldsymbol{\xi}(\boldsymbol{s}^*))=\boldsymbol{k}^*(\boldsymbol{K}+{\lambda_1}\boldsymbol{\Sigma})^{-1}\boldsymbol{\mu}
\end{equation}
\begin{equation}\label{eq:klcov}
\mathbb{D}(\boldsymbol{\xi}(\boldsymbol{s}^*))=\frac{N}{\lambda_{2}}(\boldsymbol{k}(\boldsymbol{s}^*,\boldsymbol{s}^*)-\boldsymbol{k}^*(\boldsymbol{K}+\lambda_{2}\boldsymbol{\Sigma})^{-1}\boldsymbol{k}^{*T}),
\end{equation}
where $\lambda_1>0$ and $\lambda_2>0$ are regularization factors, $\boldsymbol{k}^*\in\mathbb{R}^{B\mathcal{O}\times N\mathcal{O}}$ is a $1 \times N$ block matrix, where the $i$-th column element is $\boldsymbol{k}(\boldsymbol{s}^*,\boldsymbol{s}_i)\boldsymbol{I}_\mathcal{O}$. $\boldsymbol{K}\in\mathbb{R}^{N\mathcal{O}\times N\mathcal{O}}$ is a $N\times N$ block matrix, where the $i$-th row and $j$-th column item is $\boldsymbol{k}(\boldsymbol{s}_i,\boldsymbol{s}_j)\boldsymbol{I}_\mathcal{O}$. Besides, $\boldsymbol{{\mu}}=[\boldsymbol{\hat{\mu}}_1^T \boldsymbol{\hat{\mu}}_2^T...\boldsymbol{\hat{\mu}}_N^T]^T$ and $\boldsymbol{{\Sigma}}=blockdiag\left \{ {\hat{\boldsymbol{\Sigma}}_1},{\hat{\boldsymbol{\Sigma}}_2},...,{\hat{\boldsymbol{\Sigma}}_N} 
\right \}$.

The catching point ($t_c,\bm{x}_c,\dot{\bm{x}}_c$) generated by the QP (section \ref{subsec:QP_formulation}) is inserted into the $\mathcal{P}_n^{ref}$ as the initial point. Then, the KMP updates the desired trajectory for POC by \eqref{eq:klmean} and \eqref{eq:klcov}. \textcolor{black}{The proposed trajectory updating shares some similarities with the reference spreading control \cite{rsSaccon} since both of them modify the trajectory due to the impact}.

\subsubsection{Human Arm Stiffness Estimation}\label{subsec:human-arm}

\begin{figure}[t]
    \centering
    \includegraphics[trim=11.0cm 7.50cm 12.5cm 6.5cm,clip,width=.7\columnwidth]{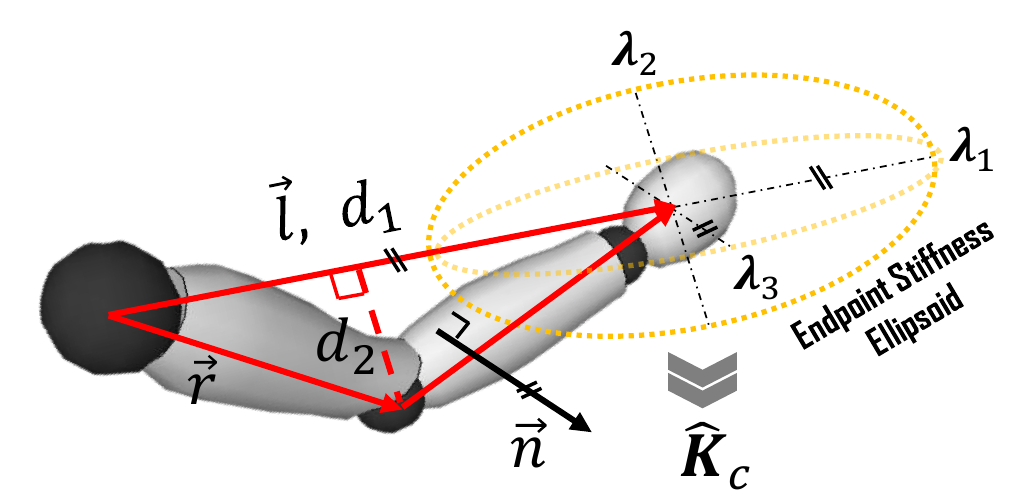} 
    \caption{{The geometric model of the human arm endpoint stiffness ellipsoid is constructed based on the arm configuration.} }
    \label{fig:armstiff}
    \vspace{-5mm}
\end{figure}

The formulation proposed in~\cite{wu2020intuitive} estimates human arm stiffness during POC. 
Fig.~\ref{fig:armstiff} presents the two-segmented human arm skeleton in 3D space, where the hand–forearm and upper arm segments compose a triangle at any non-singular configuration. The vector from the center of the shoulder joint to the position of the hand ($\vec{\boldsymbol{l}}\in\mathbb{R}^{3}$) represents the major principal direction of the human arm endpoint
stiffness ellipsoid. $\vec{\boldsymbol{r}}\in\mathbb{R}^{3}$ represents the vector from the center of the shoulder to the center of the elbow. Then, the minor principal axis direction ($\vec{\boldsymbol{n}}\in\mathbb{R}^{3}$) is defined to be perpendicular to the arm triangle plane. The remaining principal axis of the stiffness ellipsoid ($\vec{\boldsymbol{m}}\in\mathbb{R}^{3}$), which lies on the arm triangle plane, is calculated based on the orthogonality of the three principal axes. 
Hence, the orthonormal matrix $\boldsymbol{V}\in\mathbb{R}^{3 \times 3}$ can be constructed as:
\begin{equation}
	\boldsymbol{V} = \Bigg [\frac{\vec{\boldsymbol{l}}}{\parallel \vec{\boldsymbol{l}} \parallel}, \frac{(\vec{\boldsymbol{r}} \times \vec{\boldsymbol{l}}) \times \vec{\boldsymbol{l}}}{\parallel (\vec{\boldsymbol{r}} \times \vec{\boldsymbol{l}}) \times \vec{\boldsymbol{l}} \parallel}, \frac{\vec{\boldsymbol{r}} \times \vec{\boldsymbol{l}}}{\parallel \vec{\boldsymbol{r}} \times \vec{\boldsymbol{l}} \parallel} \Bigg ].
	\label{eq:eigen_vectors}
\end{equation}

The length ratio of the median principal axis to the major principal axis of the stiffness ellipsoid is inversely proportional to the distance $d_1 \in\mathbb{R}^{+}$ from the hand position to the center of the shoulder. Meanwhile, the ratio of the length of the minor principal axis to the major principal axis is assumed to be proportional to the distance $d_2 \in\mathbb{R}^{+}$ from the center of the elbow to the major principal axis.
Here, $\frac{\lambda_{2}}{\lambda_{1}} = \frac{\alpha_1}{d_1}$ and $\frac{\lambda_{3}}{\lambda_{1}} = \alpha_2 \cdot d_2$, where $\lambda_{1} \in\mathbb{R}^{+}$, $\lambda_{2} \in\mathbb{R}^{+}$ and $\lambda_{3} \in\mathbb{R}^{+}$ represent the eigenvalues corresponding to the major, median and minor principal axes, respectively. $\alpha_1 \in\mathbb{R}$ and $\alpha_2 \in\mathbb{R}$ are scalar variables, and $d_1 = \parallel \vec{\boldsymbol{l}} \parallel$, $	d_2 = \vec{\boldsymbol{r}} \cdot \frac{\vec{\boldsymbol{m}}}{\| \vec{\boldsymbol{m}} \|}$.
In this model, synergistic muscle co-contractions are assumed to contribute to the endpoint stiffness ellipsoid's volume, expressed by the active component $A_{cc}\in\mathbb{R}$. And $A_{cc}(p) = c_1 \cdot p + c_2$ 
is a linear relation with the muscle activation level $p\in\mathbb{R}$ (measured by EMG sensors). 
The diagonal matrix $\boldsymbol{D}\in\mathbb{R}^{3 \times 3}$ is formed by the length of the principal axes, i.e., the eigenvalues,
\begin{equation}
	\boldsymbol{D} = A_{cc}(p) \cdot \boldsymbol{D}_s = A_{cc}(p) \cdot 
	\frac{\text{diag} (1, \quad \alpha_1 / d_1, \quad \alpha_2 d_2)}{ (1 \times \alpha_1 / d_1 \times \alpha_2 d_2)^{\frac{1}{3}}},
\label{eq:eigen_values}
\end{equation}
where $\lambda_{1}$ is set to $A_{cc}$ , and $\boldsymbol{D}_s \in\mathbb{R}^{3 \times 3}$ represents the shape of the stiffness ellipsoid. Finally, the estimated endpoint stiffness matrix $\boldsymbol{\hat{K}}_c\in\mathbb{R}^{3 \times 3}$ is formulated by
\begin{equation}
	\boldsymbol{\hat{K}}_c = \boldsymbol{V}\boldsymbol{D}\boldsymbol{V}^T = \boldsymbol{V}A_{cc}(p)\boldsymbol{D}_s\boldsymbol{V}^T.
	\label{eq:stiffness_model}
\end{equation}
We use the average value of the four parameters identified by \cite{wu2020intuitive}: \resizebox{.9\columnwidth}{!}{$c_1=2033.325 $, $c_2=140.606 $, $\alpha_1=0.255$, $\alpha_2=2.815$}. The Cholesky decomposition is used to decompose the $\boldsymbol{\hat{K}}_c$, encoded by the GMM/GMR later, as described in~\cite{abu2018forceras}, which will output the robot's desired stiffness $\boldsymbol{K}^{d}$.

\subsection{Cartesian Impedance Controller}\label{subsec:impedance_controller}

A Cartesian Impedance Controller is used to deal with the free and constrained motions \cite{Zhao2022}. Since only stiffness is estimated from the demonstrations, as discussed in \ref{subsec:human-arm}, we use a double diagonal method to obtain the desired damping value $\boldsymbol{D}^{d}$ as a function of $\boldsymbol{K}^{d}$. The VIC at each time step generates a new stiffness and, consequently, a new damping value
as $diag(\boldsymbol{D}^d(t)) = 2 \zeta \sqrt{diag(\boldsymbol{K}^d(t))}$ where $diag(\cdot)$ is the matrix diagonal components at time $t$, and the damping factor $\zeta=0.707$ \cite{ott2008cartesian}. Note that a high constant stiffness is used until the impact force triggers the learned human variable stiffness (HVS).

\section{Experiments and Results}

\subsection{Human Catching Demonstrations Collection}\label{subsec:human-demo}

An experimental investigation was performed on a healthy human subject to collect and analyze the human-compliant behavior in POC (see Fig.~\ref{fig:con_fig} bottom left side).

\subsubsection{Setup}

The subject was required to wear a Lycra suit covered with markers, tracked online by the OptiTrack to collect the human upper-body motion data. 
An EMG system was employed to measure muscle activity.
The data acquisition and synchronization of all the sensors are managed using Robot Operating System (ROS) environment at 
50 Hz. 

\subsubsection{Protocol} 
The subject was required to perform $9$ catching tasks according to the following protocol: at the beginning of each trial, the subject was told to maintain his feet steady, with the arms placed forward at ninety degrees to the hips, waiting for the box. An experiment assistant, located in front of the subject, released a $5$ kg box at a fixed height of $1$ m from the subject's arms. 
The weight of the box remained unknown to the subject to obtain the most natural behavior~\cite{humanCatching2021}.

\begin{figure}
    \centering
    \includegraphics[width=0.9\linewidth]{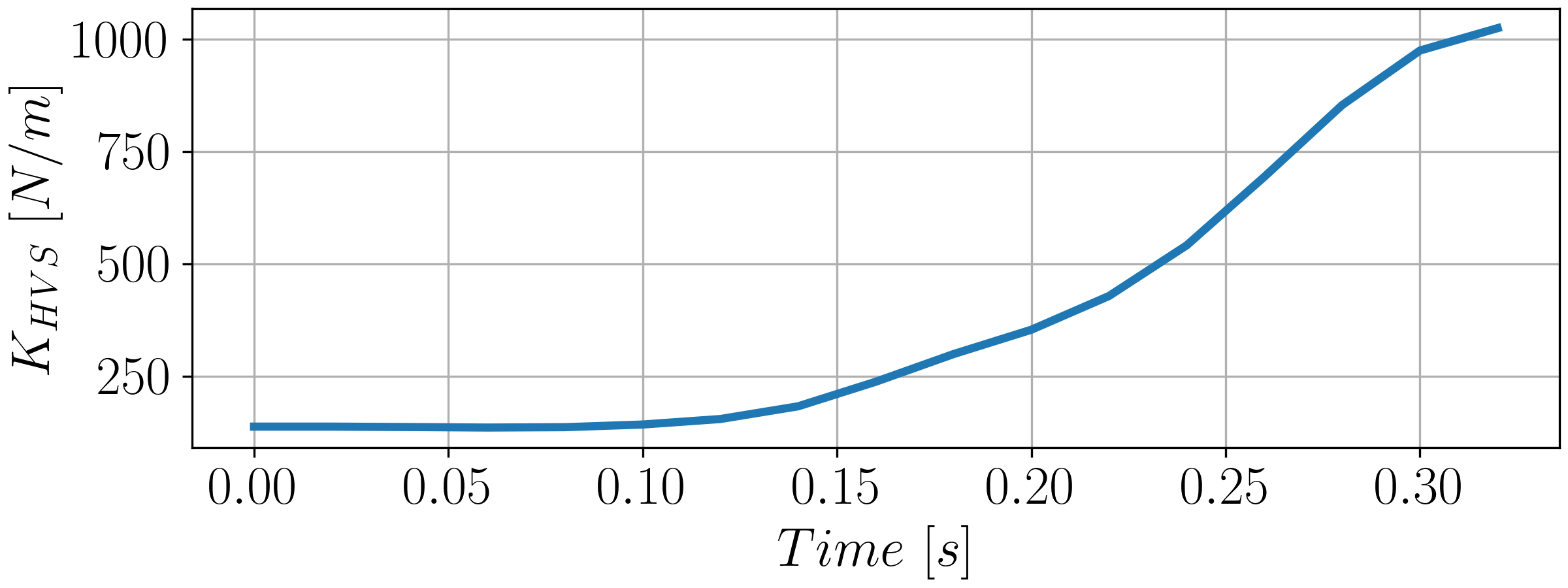}
    \caption{Human arm stiffness profile learned from demonstrations. It corresponds from the moment the human touches the object until the lowest point in the trajectory.}
    \label{fig:hvs}
    \vspace{-1.5mm}
\end{figure}

Finally, $4$ demonstrations were selected, and the mean value of both arms was used to construct the stiffness (see Section \ref{subsec:human-arm}). Only the biceps brachii was analyzed since it is the most activated muscle in the catching task. The learned HVS are presented in Fig. \ref{fig:hvs}. Although the HVS is encoded w.r.t the relative distance, we present the arm stiffness through time so it is possible to compare the profile in the results section.

\begin{figure} [t]
    \centering
    \includegraphics[width=0.7\linewidth]{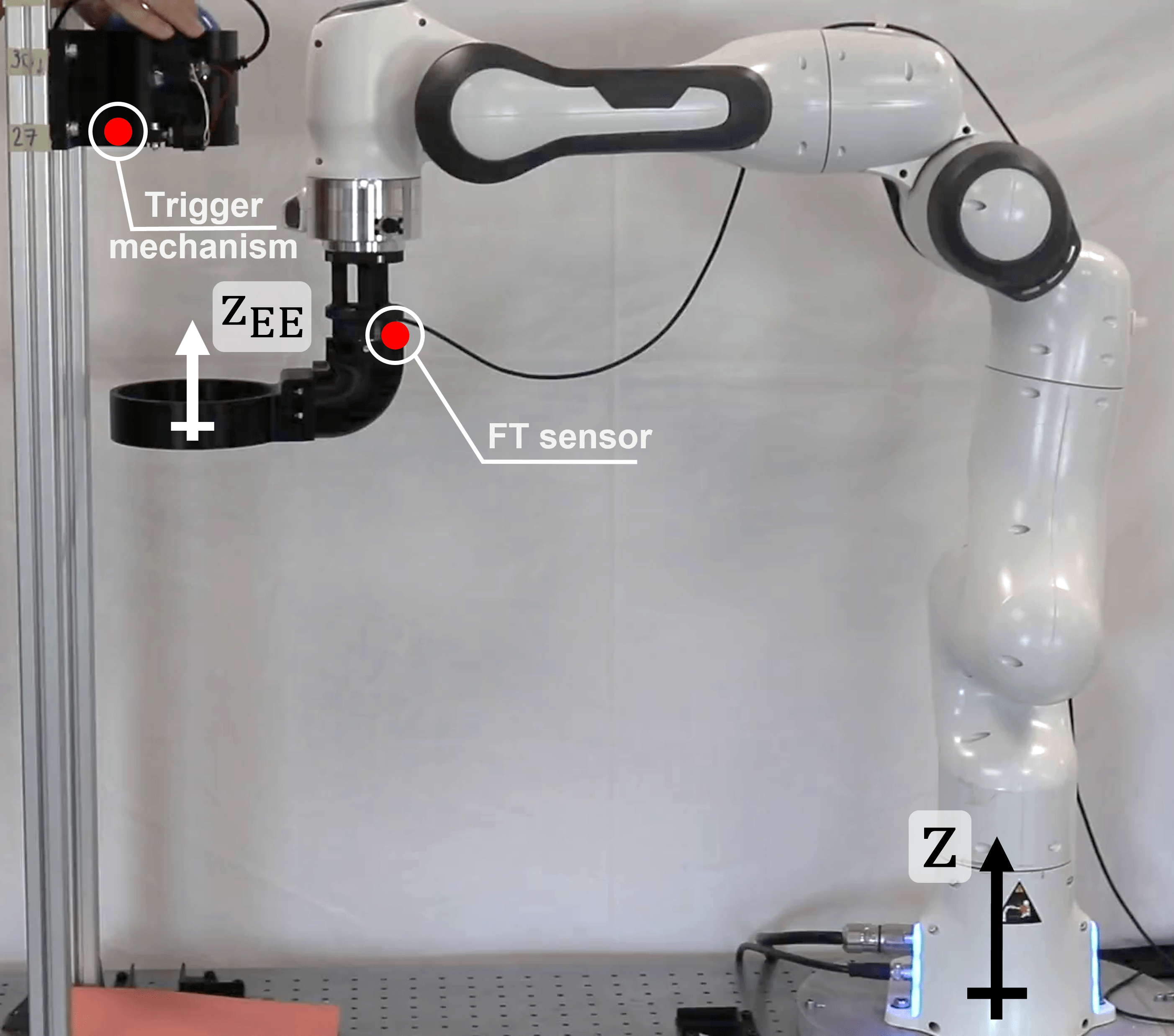}
    \caption{Experimental setup: the tool is connected to the robot through a force-torque sensor (ATI Mini45), with a combined mass of $0.47$ kg; the world z-axis is aligned with the z-axis of the EE. The sensor was added since estimated forces and torques are inaccurate in impact studies.}
    \label{fig:exp_setup}
    \vspace{-5mm}
\end{figure}

\subsection{Experimental Setup}\label{subsec:exp-set}

\begin{figure*}[ht]
    \centering
    \includegraphics[trim=28.27cm 0.0cm 27cm 5.5cm,clip,width=0.28\columnwidth]{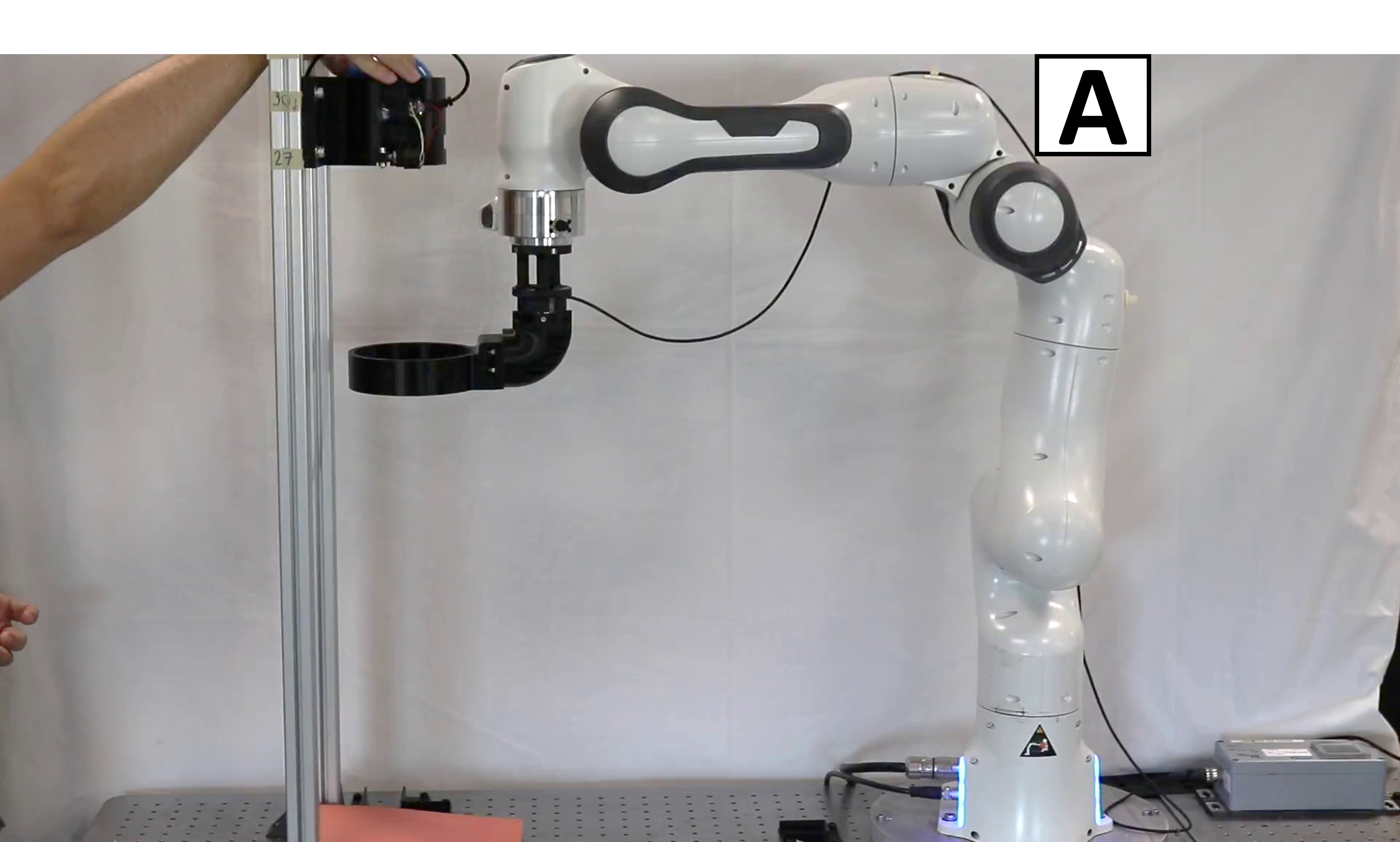}
    \includegraphics[trim=28.27cm 0.0cm 27cm 5.5cm,clip,width=0.28\columnwidth]{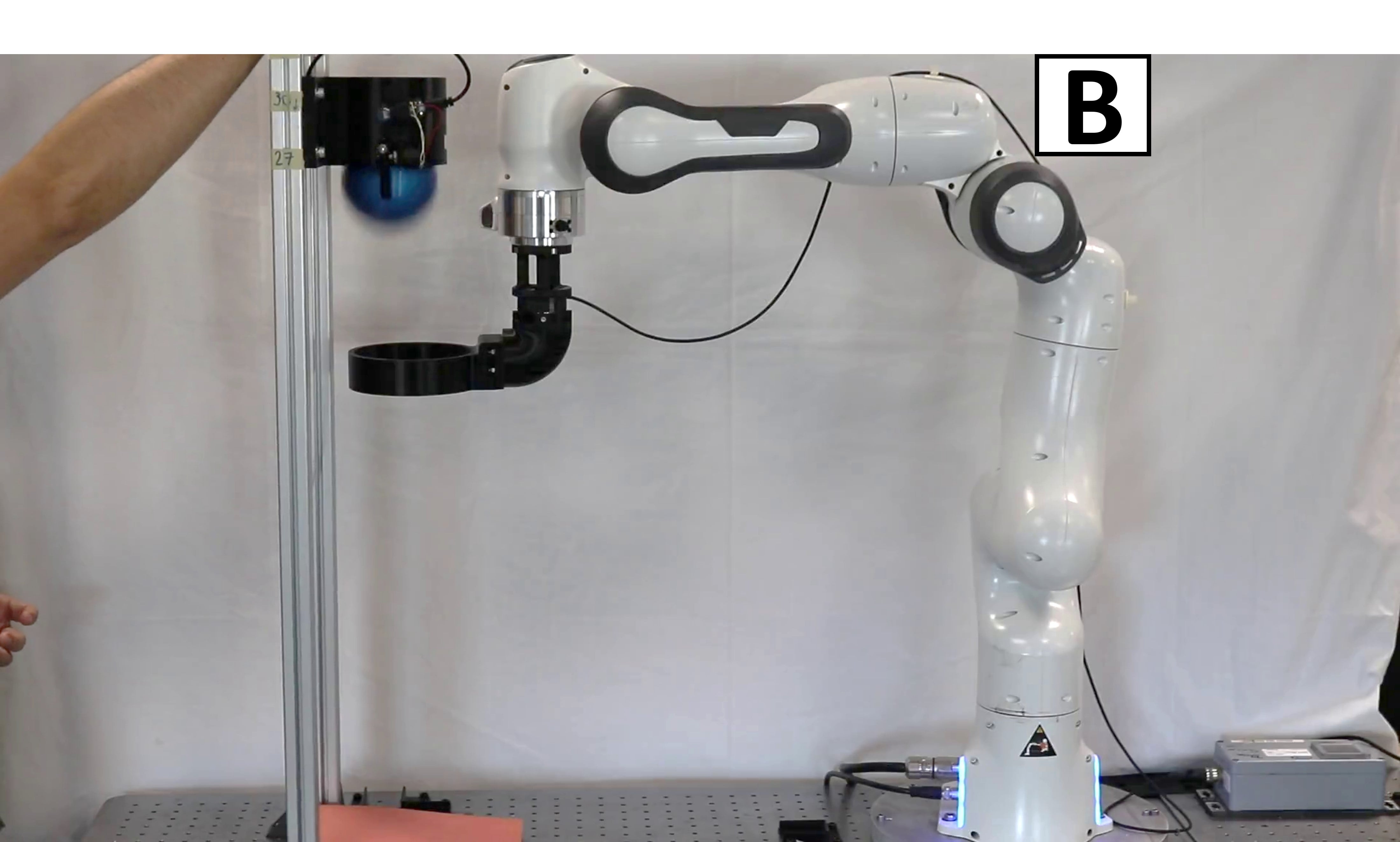}
    \includegraphics[trim=28.27cm 0.0cm 27cm 5.5cm,clip,width=0.28\columnwidth]{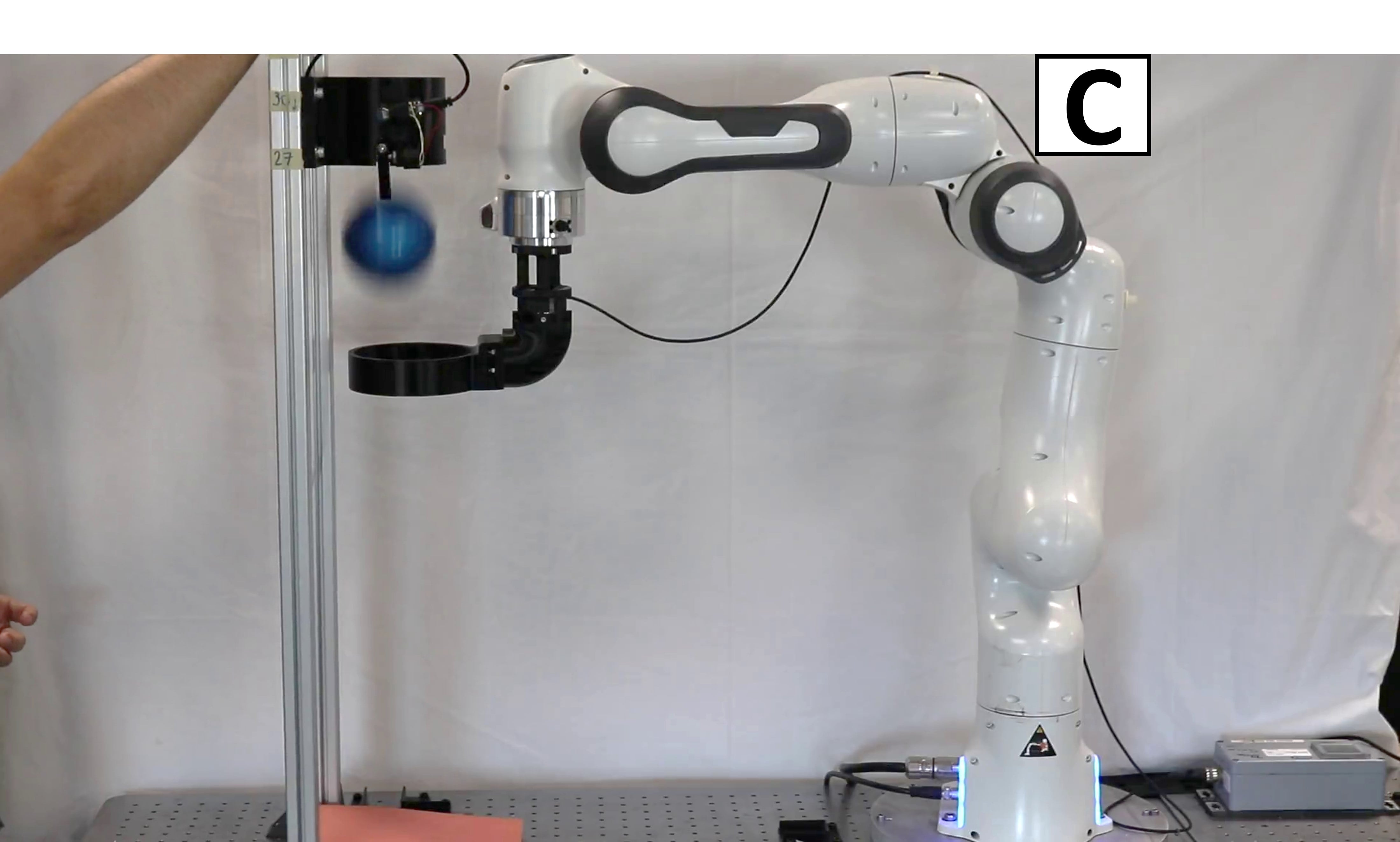}
    \includegraphics[trim=28.27cm 0.0cm 27cm 5.5cm,clip,width=0.28\columnwidth]{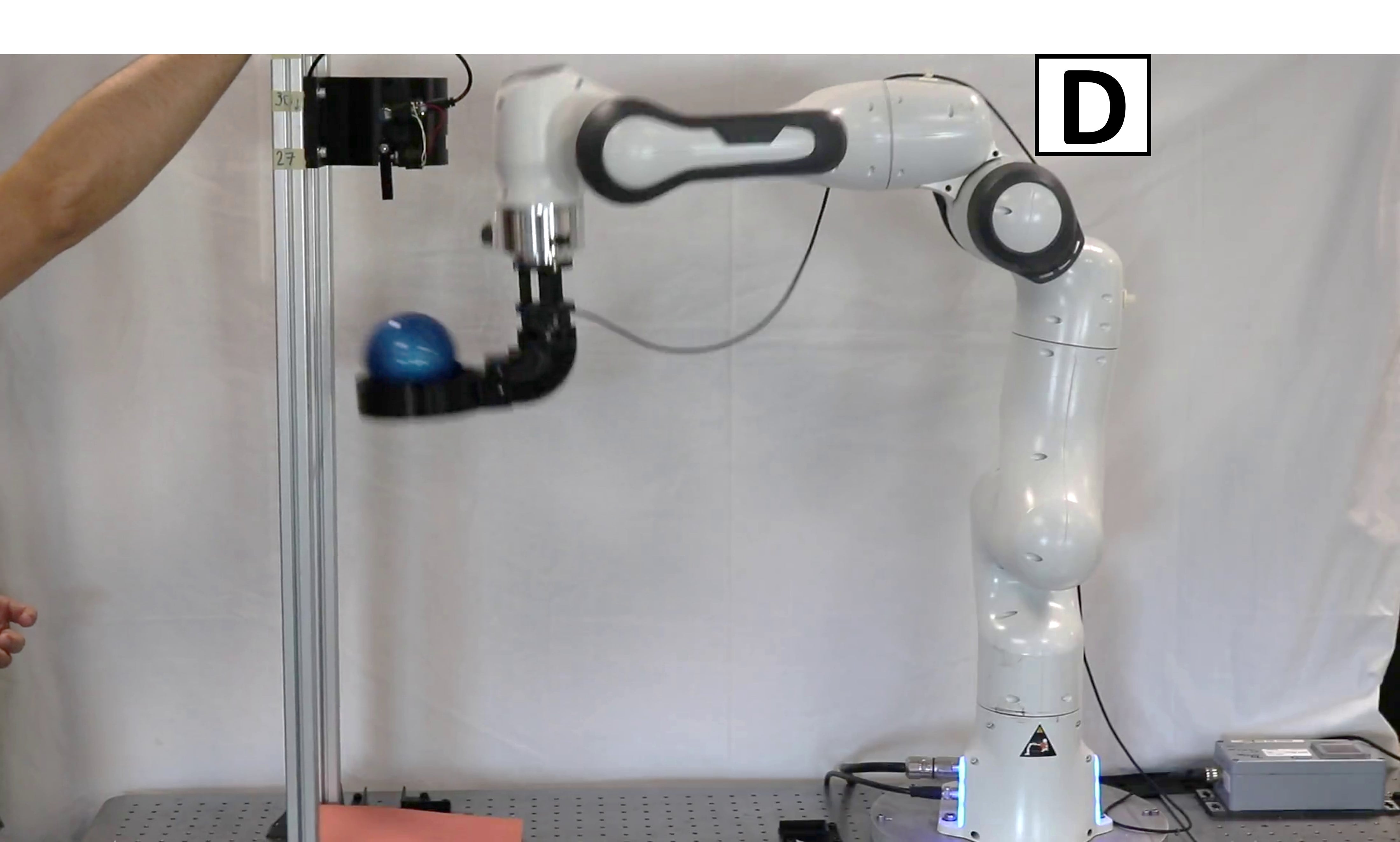}
    \includegraphics[trim=28.27cm 0.0cm 27cm 5.5cm,clip,width=0.28\columnwidth]{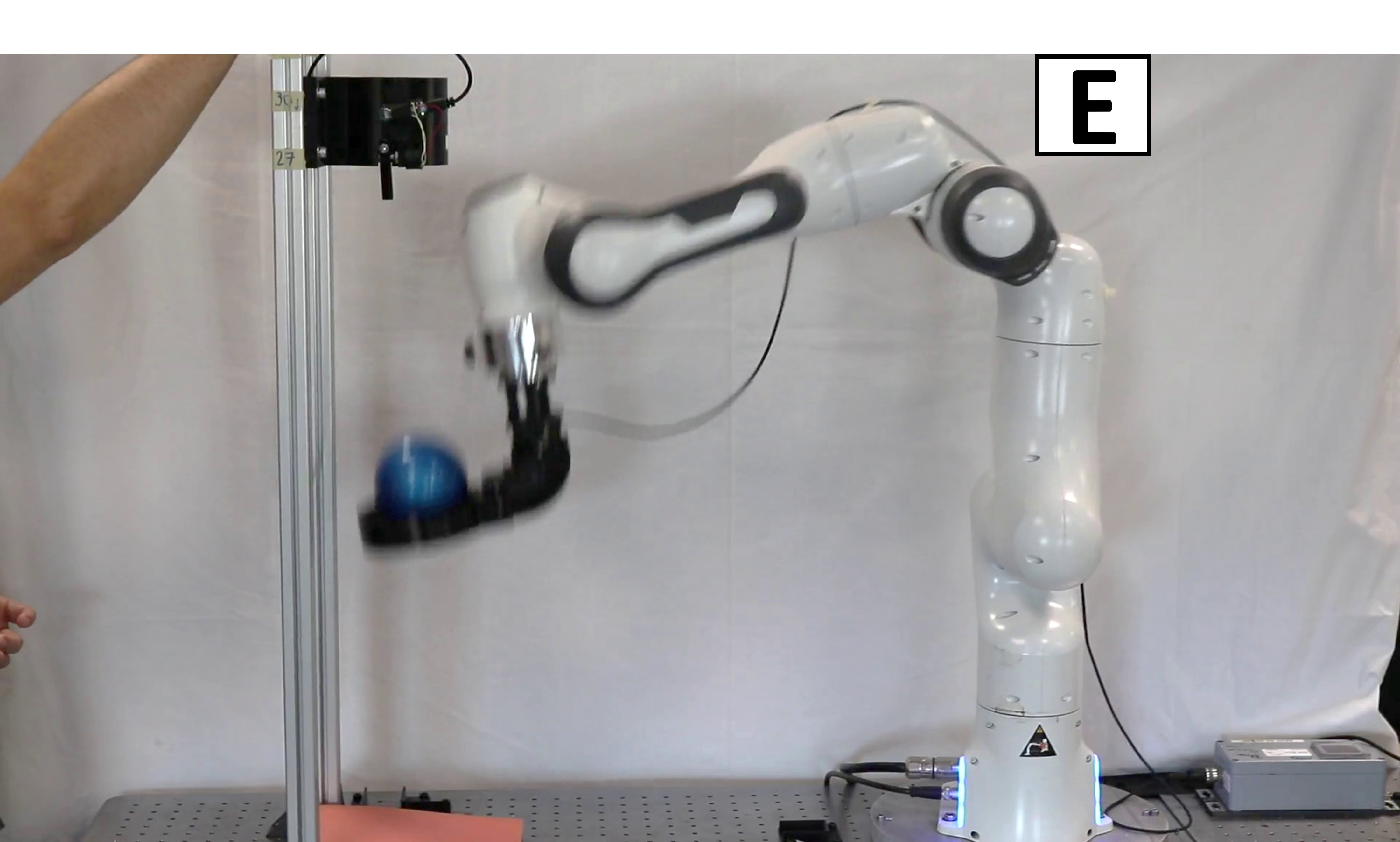}
    \includegraphics[trim=28.27cm 0.0cm 27cm 5.5cm,clip,width=0.28\columnwidth]{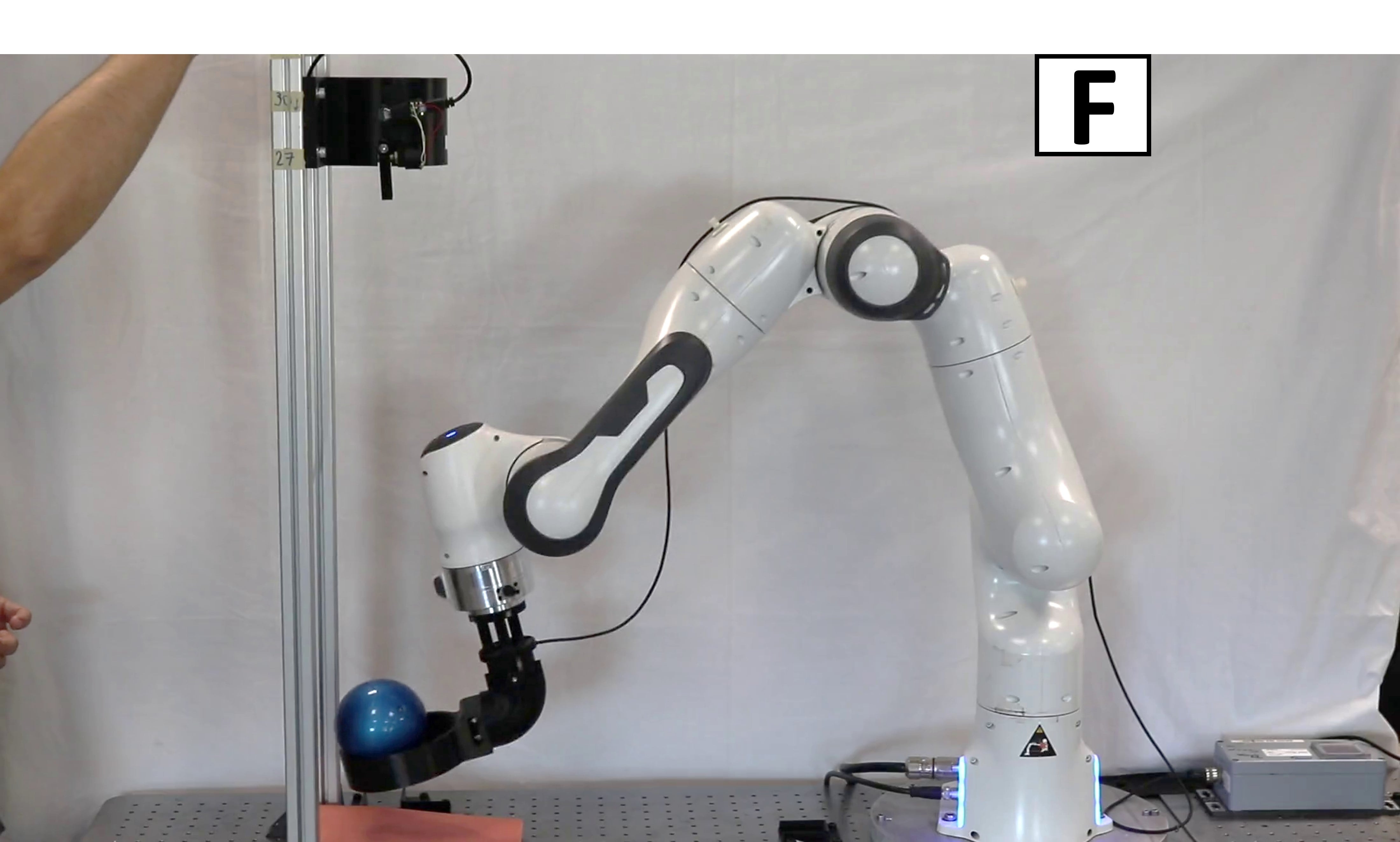}
    \includegraphics[trim=28.27cm 0.0cm 27cm 5.5cm,clip,width=0.28\columnwidth]{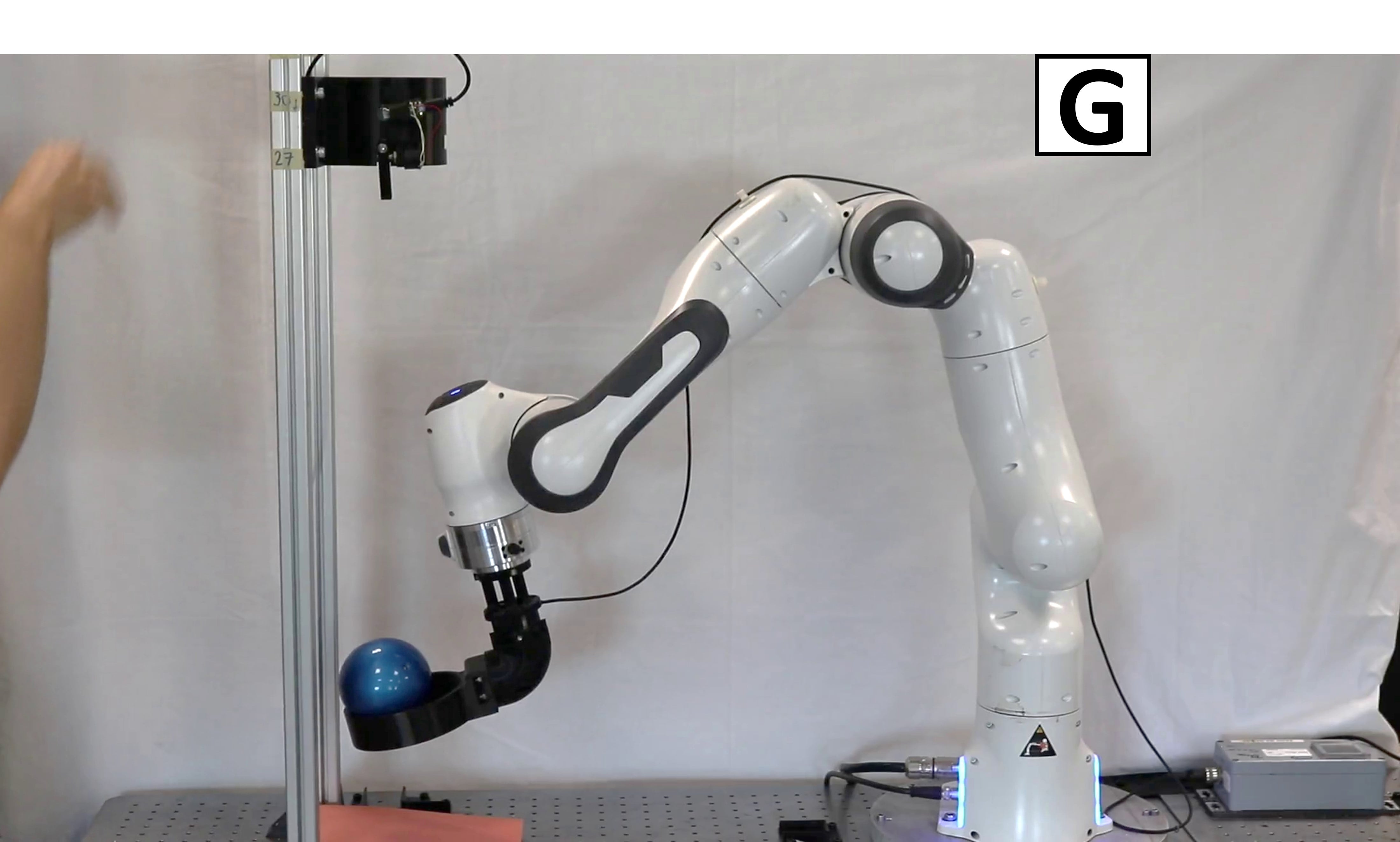}
    \caption{A time sequence of the catching task: PRC phase is depicted in frames A to C, while D to G shows the POC. The impact is shown in frame D, and frame F shows the lower point in the trajectory.}
    \label{fig:frames_catching}
    \vspace{-6 mm}
\end{figure*}

\begin{figure*}[ht]
    \subfloat[]{
        \centering
        \includegraphics[width=.48\textwidth, trim={0.6cm 0.9cm 0.8cm 0.5cm},clip]{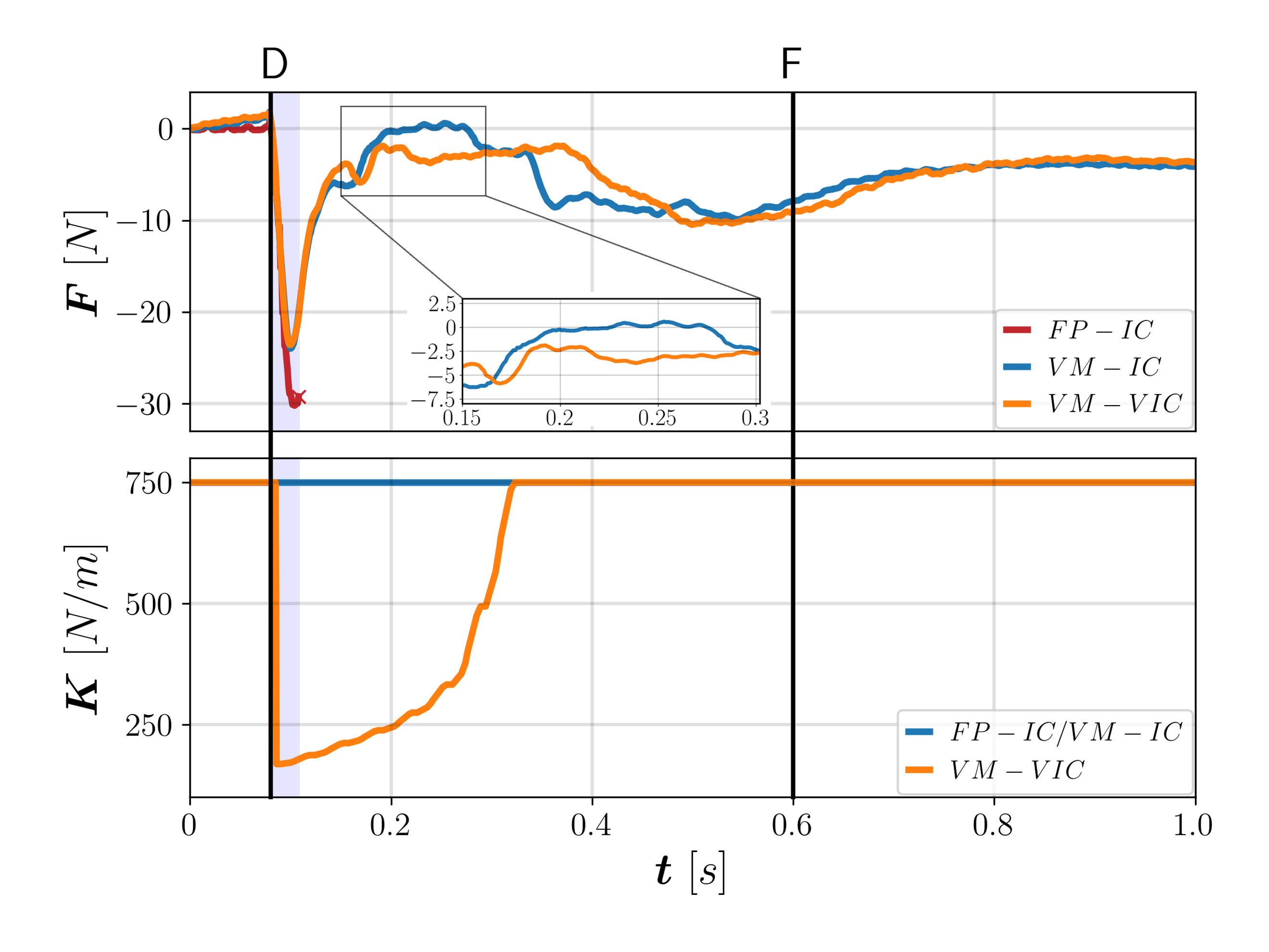} \label{fig:time_plots_force_impedance}
        } 
    \hfill
    \subfloat[]{
        \centering
        \includegraphics[width=.48\textwidth,trim={0.6cm 0.9cm 0.8cm .5cm},clip]{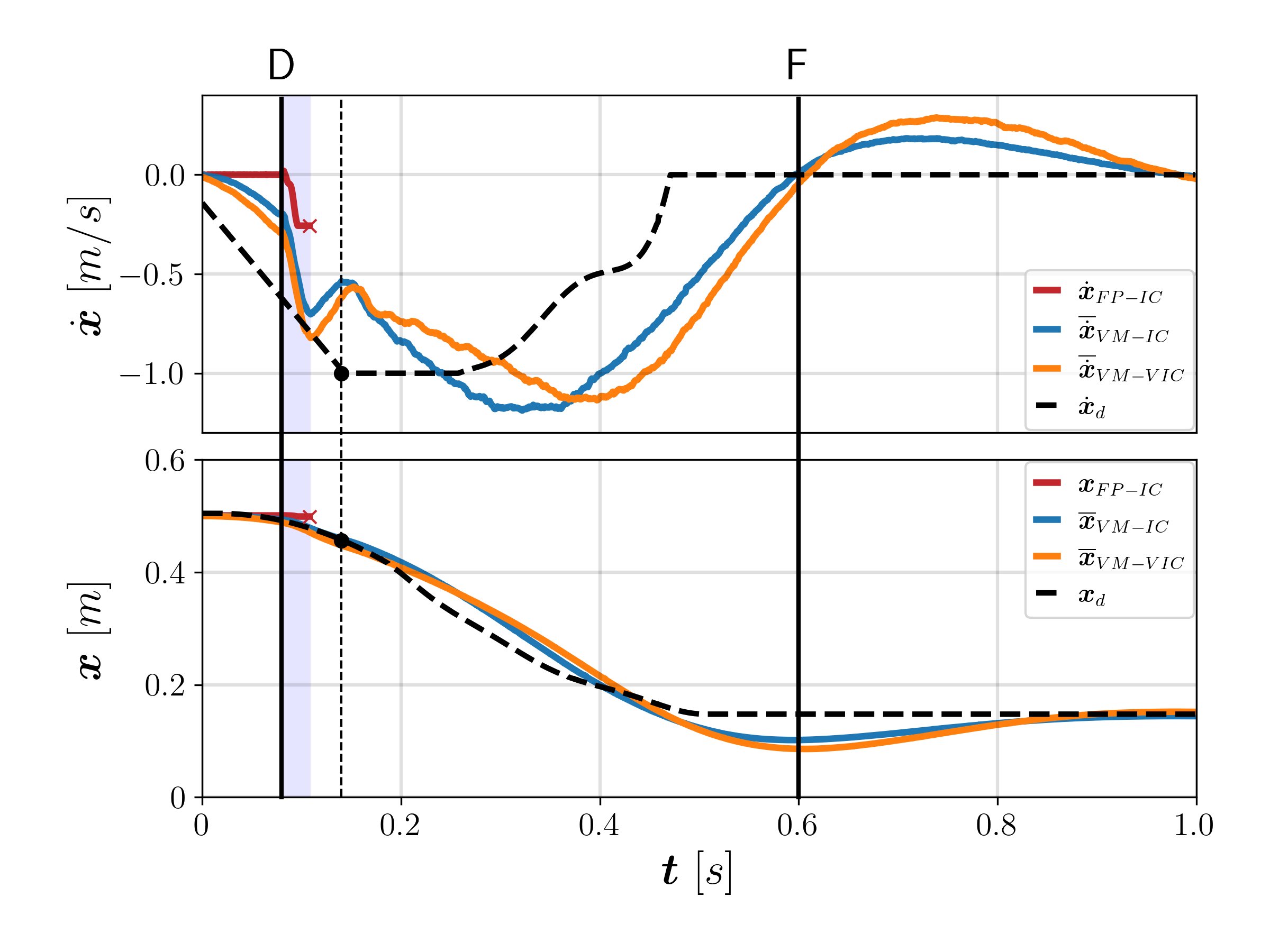} \label{fig:time_plots_twist_pose}
        }

    \caption{Temporal plots for the proposed task along the z-axis: a) interaction force (top) and stiffness (bottom), b) velocity (top) and position (bottom). Snapshot D (Fig. \ref{fig:frames_catching}) is the first impact, which starts at around 80 ms and ends at 110 ms (blue background). The dashed vertical line in (b) represents the predicted impact position and velocity. All legends with bar overline stand for average.}
    
    \label{fig:time_plots}
    \vspace{-5 mm}
\end{figure*}

The experimental setup consists of a 7-DoFs Franka Emika Panda robotic arm with a 3D-printed flat EE, useful to isolate the catching impact along the vertical axis (Fig. \ref{fig:exp_setup}). The Panda has a payload of $3$ kg suggested by the manufacturer, often designed to guarantee positioning accuracy \cite{FrankaDoc}. However, the allowed effective payload may be smaller during fast movements due to the additional inertial loads. Therefore, even if the object-tool assembly is below the maximum allowed load, the object's high velocity and acceleration might still lead to failure by triggering the safety system. The declared EE's maximum velocity is $1.7$ m/s, which was never achieved in the constrained workspace used in these experiments. Since the maximum velocity and acceleration are essential for VM planning, the maximum velocity adopted is $\dot{\boldsymbol{x}}^{max}_R = \left[ \dot{\bm{x}}^{max^T}, \hdots, \dot{\bm{x}}^{max^T} \right]^T$ with $\dot{\bm{x}}^{max} =  \left[ \dot{x}^{max}_{pos}, \dot{x}^{max}_{pos}, \dot{x}^{max}_{pos}, \dot{x}^{max}_{rot}, \dot{x}^{max}_{rot}, \dot{x}^{max}_{rot} \right]^T$ in which $ \dot{x}^{max}_{pos}=1.0$ m/s and $ \dot{x}^{max}_{rot}=2.5$ rad/s.
The same structure holds for the accelerations, with $\ddot{x}^{max}_{pos}=6.0$ m/s$^2$ and $ \ddot{x}^{max}_{rot}=25.0$ rad/s$^2$.

Since the free-falling object is supposed to execute a 1-DoF trajectory, a reference mechanism was designed to ensure a consistent initial position of the object. A triggering mechanism guaranteed synchronization between releasing the thing and the robot's movements. The trigger consists of a potentiometer with a 3D-printed part that is mechanically attached to it and works as a lever moved by the object, and an Arduino MKR1000 reads its signal. The catching object is a ball with a mass of $0.5$ kg and a non-deformed radius of $47.5$ mm. The ball falls from the same height in each experiment to facilitate methods comparison, with a relative height along the z-axis of $27$ cm\footnote{This height was chosen empirically to avoid collisions during the movement. The ball's velocity after falling $27$ cm is about $2.3$ m/s, more than twice the maximum adopted velocity of the robot. Thus, the impact is inevitable.}.
All the stiffness matrices used are $\boldsymbol{K}^d=diag(750, 750, 750)$ N/m and are the initial and final values for the VIC. These values are chosen to improve trajectory tracking performances. Since the impact force is not related to the impedance controller's stiffness \cite{haddadin2009requirements}, there is no need to evaluate a lower stiffness in the experiments. 
The QP solver is ALGLIB QP-BLEIC on Ubuntu 20.04, with $\alpha=0.15, \beta=1,\gamma=1$. All experiments run on a computer with Intel Core i7-11700 2.5 GHz $\times$ 16-cores CPU and 32 GB RAM.

\renewcommand{\arraystretch}{1.1}
\begin{table}[t] \caption{Metrics and results of the free-falling object experiments}
\centering
\begin{adjustbox}{width=.49\textwidth} \label{tab:metrics-results}
\begin{tabular}{cccccc}
          \large \textbf{LOI} [Ns]               & \large \textbf{DRI} [-]  & \large \textbf{BTI} [ms]              & \large \textbf{E} [J] & \large \textbf{F}$_{max}$ [N] \\ \hline
          \\ 
           $\int_{\Delta t'}|\boldsymbol{F}_z-\boldsymbol{F}_w| dt$    &  $(1+(\frac{2\pi}{\delta})^2)^{-0.5}$       &  $t_{\max{(\boldsymbol{F}_z,0)}}$        &  $\int_{\Delta t'} \boldsymbol{F}^T \dot{\boldsymbol{x}} dt$            &  $\max(|\boldsymbol{F}_z|)$   \\
          \\ \hline
\textbf{FP-IC}: -                                & -       & -                                   & -              & 30.2   \\
\textbf{VM-IC}:  2.5                              & 0.38        & 50                                   & 1.92          & 24.0   \\
\textbf{VM-VIC}: $\bm{2.3}$                              & $\bm{0.42}$        & $\bm{0}$                                    & $\bm{1.97}$          & $\bm{23.6}$   \\ \hline
\end{tabular}
\end{adjustbox}
\vspace{-4.5 mm}
\end{table}

\subsection{Metrics}

To analyze the performance of the proposed methods precisely, we use the quantitative metrics proposed in \cite{ajoudani2012tele} and shown in table \ref{tab:metrics-results}.
The `lift off index' (LOI) represents the integral of the difference between the vertical component of the wrist force $\boldsymbol{F}_z$ and its steady-state value (corresponding to the weight of the ball $\boldsymbol{F}_w$) during $\Delta t'$, i.e., from impact until steady state. A higher LOI indicates multiple bouncing and/or long under-damped ball trajectories.
The `damping ratio index' (DRI) is the logarithmic decrement between the first and second impact force peaks $f_{z_{p,1}},f_{z_{p,2}}$, respectively, with $\delta=\log ( \frac{f_{z_{p,1}}}{f_{z_{p,2}}} )$, which indicates the capability to absorb impact force and energy and damp it quickly.
Finally, the `bouncing time index' (BTI) is the period when the contact between the ball and the robot is lost. Dissipated energy ($\boldsymbol{E}$) and maximum force ($\boldsymbol{F}_{max}$) are also taken into account.

\subsection{Experimental Results and Discussions}

We evaluated three different cases: a fixed-position impedance controller (FP-IC) with constant impedance, used as baseline; a VM algorithm with impedance controller (VM-IC) that uses QP+KMP as planners but without changing the impedance; the last one use the same planning strategy as VM-IC, i.e., QP+KMP for VM, and adds the VIC for the post-impact stage (VM-VIC). 
A snapshot sequence with VM is shown in Fig. \ref{fig:frames_catching}. The PRC and POC stages are shown by the frames A-B-C and D-E-F-G, respectively. The FP-IC baseline keeps the pose shown in frame A as the desired position for the catching. Fig. \ref{fig:time_plots} shows all experiments' interaction forces, positions, velocities, and stiffnesses through time. VM-IC and VM-VIC are averages of three repetitions, given the strong repeatability and similarity among trials, while FP-IC is the result of one experiment since it triggered the robot's safety system, as discussed in the next section. 
\subsubsection*{Force and impedance analysis}
the FP-IC was not able to conclude the task. The Panda's safety system was triggered, the brakes were activated, and signal acquisition stopped. Fig. \ref{fig:time_plots} shows a red cross at the moment the robot stops. The results for the FP-IC in table \ref{tab:metrics-results} only show the maximum force since it occurs when the robot is still functioning.

VM-IC and VM-VIC completed the task and had similar behavior during impact due to the same PRC's VM profile. 
Impact occurred between 80 ms and 110 ms (blue region in Fig. \ref{fig:time_plots}). The maximum force occurs during this time frame, where FP-IC reached $30.2$ N, approximately 26\% more when compared to VM-IC ($24.0$ N) and 28\% compared to VM-VIC ($23.6$ N). This confirms the importance of VM in impact-friendly controllers for catching since the impact is faster than the controller-mechanical system response time.

The VM-IC has a lesser damping capability when compared to the VM-VIC, and after $t>110$ ms, it is possible to notice the differences. First, after the impact, the VM-VIC never loses contact with the object, as shown in the detail of Fig. \ref{fig:time_plots_force_impedance} (top), i.e., $F_z<0$ at all time steps within $t>110$ ms (BTI $=0$). Meanwhile, the VM-IC reaches forces above zero for 50 ms (BTI $>0$). Bouncing may lead to several problems when catching objects with interaction controllers, and a more thorough discussion may be found in \cite{ajoudani2012tele}. Second, since BTI happens in a very narrow time window, we reinforce the lesser damping capability statement by checking the LOI: VM-IC has $2.5$ Ns against VM-VIC's $2.3$ Ns, indicating a less damped behavior for VM-IC. Third, DRI is smaller for VM-IC ($0.38$) than for VM-VIC ($0.42$): the first force peak is during the impact, where VM plays a more predominant role; in contrast, the second peak happens during manipulation, confirming the importance of adaptive control after impact. Finally, the extracted energy using VM-VIC (E $=1.97$ J) is higher than VM-IC (E $=1.92$ J), indicating a higher dissipation from VM-VIC.

Regarding impedance adaptation, the profile in Fig. \ref{fig:time_plots_force_impedance} (bottom) shows the stiffness learned with LfD taking effect. It is triggered with a force threshold of 3 N. The stiffness is immediately reduced to absorb the energy when the impact happens, and it is then gradually increased until the maximum value, observing the HVS learned. The learned HVS value is set to the empirically defined maximum of 750 N/m, when it is above that threshold. \textcolor{black}{Although the experimental setup here is different from the human demonstrations, the learned HVS still outperformance the constant stiffness, showing the proposed method's adaptability.}

\begin{figure}
    \centering
    \includegraphics[width=.9\linewidth]{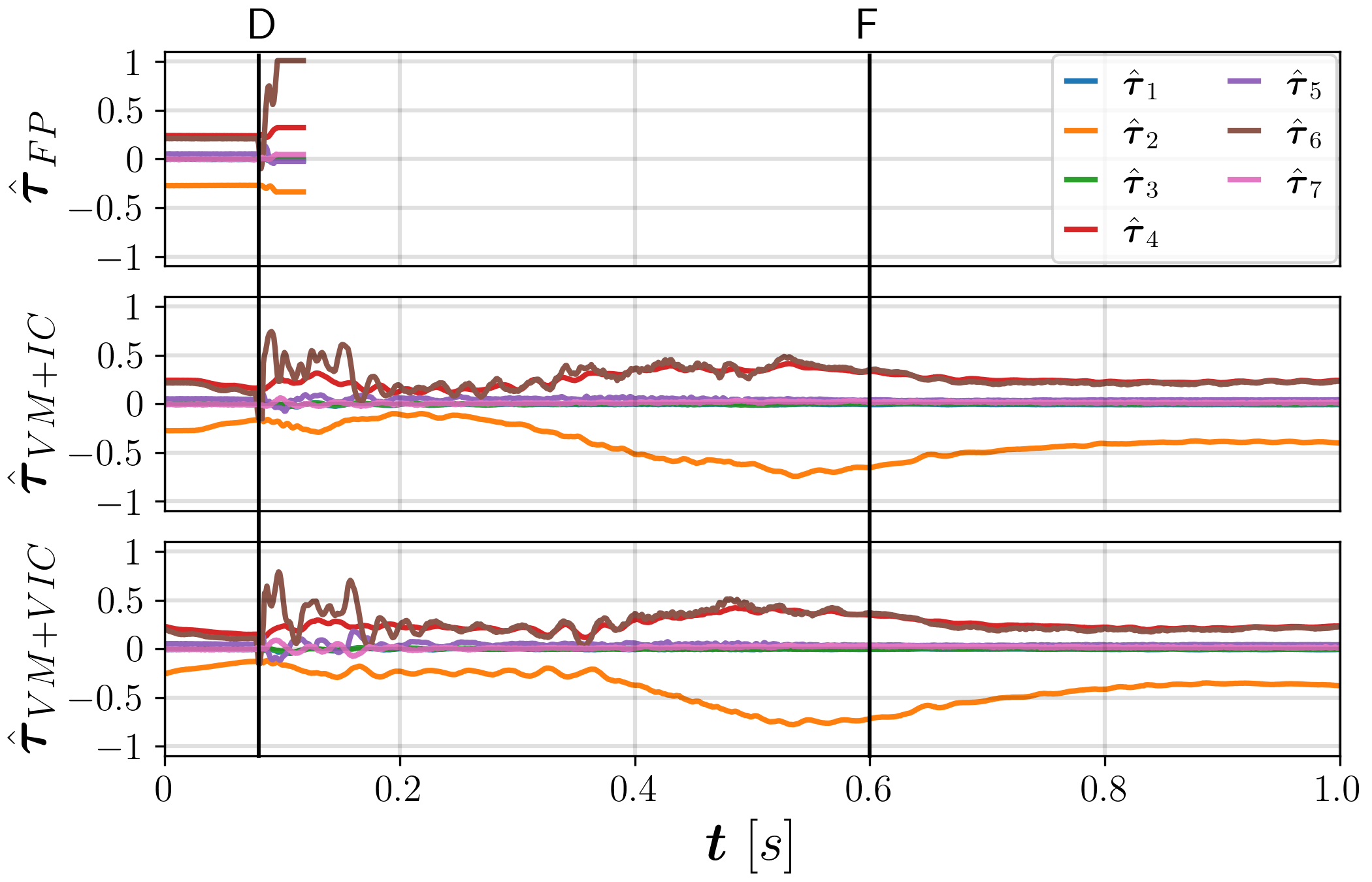}
    \caption{Normalized joint torques measured during each experiment: FP-IC (top), VM-IC (middle), and VM-VIC (bottom).}
    \label{fig:torques}
    \vspace{-1.5mm}
\end{figure}

\begin{figure}
    \centering
    \includegraphics[width=0.9\linewidth]{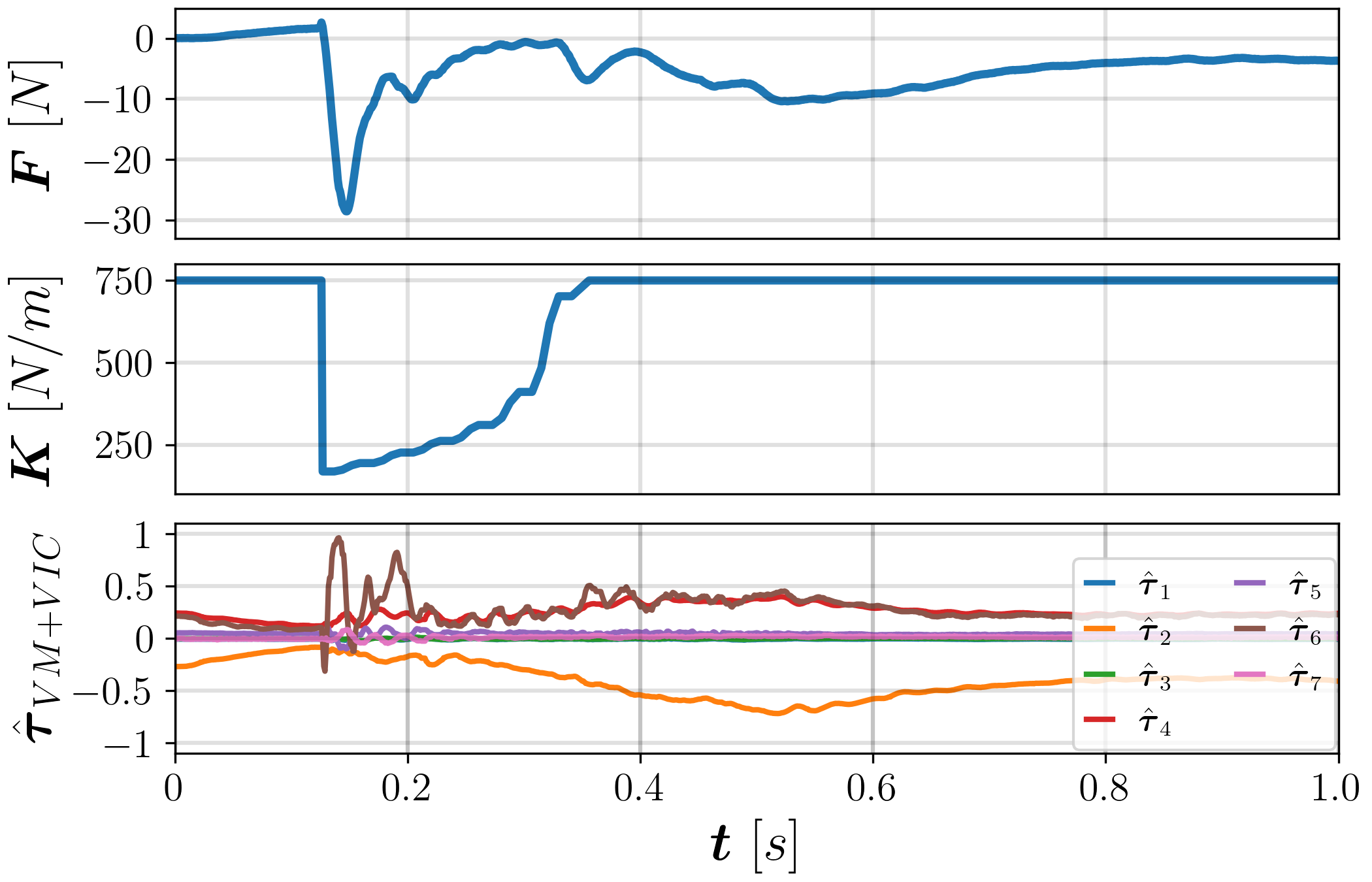}
    \caption{Experimental results for the new dropping height of 37 cm: measured force along $z-$axis (top), Cartesian stiffness (middle), and joint torques (bottom).}
    \label{fig:exp-37cm}
    \vspace{-4.5mm}
\end{figure}

\subsubsection*{Position and velocity analysis}
the dashed vertical line in Fig. \ref{fig:time_plots_twist_pose} is the predicted catching point, after which the position and velocity desired profiles are learned from the human (the desired velocity is constrained to the robot's limit). Since the ball and the robot are triggered simultaneously, and the object is already inside the robot's workspace, the inherent delay in position and velocity tracking from their desired values brings the impact earlier than predicted. In tasks where the object comes from outside the robot's workspace, the robot would have more time to react,
therefore improving VM. However, this may be dangerous for the equipment with objects having a reasonably high mass. 

\subsubsection*{Joint torque analysis} the reason why the robot blocks in the FP-IC experiments, is the excessive reflected joint torques due to the impact. Some joints may reach values above the safety limit during the experiment \cite{FrankaDoc}, triggering the safety system and aborting the experiment. The joint torques are shown in Fig. \ref{fig:torques}, and, due to different hardware in each joint, each torque is normalized according to the maximum allowed torque at that joint, thus, $\hat{\tau}_i=\tau_i/\tau^{max}_i,~i\in[1,7]$. Joint six is overloaded in the FP case, always stopping the experiment. On the other hand, the proposed controllers VM-IC and VM-VIC, remained below the safety limits. 

Despite the peak torques related to joint six, it is also visible how most of the load acting on the robot is absorbed by joint number 2, which has the overall highest activation throughout the task, at the expense of most of the other joints. This is mainly related to the robot's configuration and will be further addressed in future works by exploiting the redundant configuration (via reflective mass minimization) to optimize the internal torques' distribution.

\subsection{Experimental generalization: increased height} 

An extension of the experiment is studied to evaluate the proposed method's generalization, by increasing the initial height of the object by ten centimeters, i.e., with a relative total height of 37 cm. Fig. \ref{fig:exp-37cm} shows the results for the VM-VIC method. Even though the increase in height leads to a higher impact, the peak force ($|\boldsymbol{F}|_{max}=28.5$ N) is still below the value obtained with the FP-IC at 27 cm, thus a successful catch is accomplished. No bouncing is detected (BTI $=0$), but the method's behavior is slightly less damped (DRI $=0.14$). Although the successful catch, the torque at joint six almost reached the limit, indicating that the experimental setup is close to the robot's limits. Therefore, the proposed method increased the workspace range for the catching task.

\section{Conclusions}
This paper aimed to address the issue of non-prehensile catching of \textcolor{black}{falling} objects by minimizing impact-related force exchanges.
We have formulated a complete framework for generating pre- and post-impact robot trajectories and stiffness profiles. In particular, the trajectories up to the optimal catching point are generated via QP optimization for maximum VM, whereas the variable stiffness and post-impact trajectories are learned from human demonstrations.
As shown, with the baseline constant impedance controller, it was impossible to complete the experiments, given the excessive impact forces, resulting in the motors' emergency lock. 
However, the proposed method reduces the instantaneous impact forces exchanged. The VIC achieves a human-like motion in the POC phase, improving energy absorption and other performance metrics. Moreover, the method allowed the robot to catch an object within a higher dropping position, increasing the workspace for the proposed task. Although the robot reached high joint torques, a further evaluation of methods to avoid joint torque overload can be considered.
Future works will encompass a more complex configuration-dependent controller to fully exploit the robot's redundancy, improve the velocity tracking, and minimize internal torques. 

\bibliographystyle{IEEEtran}
\bibliography{biblio}

\end{document}